\documentclass{article}

\usepackage{arxiv}

\usepackage[utf8]{inputenc} 
\usepackage[T1]{fontenc}    
\usepackage{hyperref}       
\usepackage{url}            
\usepackage{booktabs}       
\usepackage{amsfonts}       
\usepackage{nicefrac}       
\usepackage{microtype}      
\usepackage{lipsum}
\usepackage{graphicx}
\graphicspath{{./images/}}
\usepackage[linesnumbered,ruled,vlined]{algorithm2e}
\usepackage{wrapfig}
\usepackage{makecell}
\usepackage{multirow}
\usepackage{pifont}
\usepackage{comment}
\usepackage{bm}
\usepackage{pifont}
\usepackage{amssymb}
\usepackage{subcaption}
\usepackage{amsthm}
\usepackage{CJK}
\usepackage{array}
\usepackage{amsmath}
\usepackage{natbib}
\usepackage{doi}
\usepackage{wrapfig}
\newtheorem{definition}{Definition}

\title{PIE: a Parameter and Inference Efficient Solution for Large Scale Knowledge Graph Embedding Reasoning}

\author{
 Linlin Chao, Xiexiong Lin, Taifeng Wang, Wei Chu \\
 Ant Group\\
  \texttt{ \{chulin.cll, xiexiong.lxx, taifeng.wang, weichu.cw\}@antgroup.com} \\
}

\begin{document}
\maketitle
\begin{abstract}
Knowledge graph (KG) embedding methods which map entities and relations to unique embeddings in the KG have shown promising results on many reasoning tasks.
However, the same embedding dimension for both dense entities and sparse entities will cause either over parameterization (sparse entities) or under fitting (dense entities). Normally, a large dimension is set to get better performance.
Meanwhile, the inference time grows log-linearly with the number of entities for all entities are traversed and compared.
Both the parameter and inference become challenges when working with huge amounts of entities.
Thus, we propose PIE, a \textbf{p}arameter and \textbf{i}nference \textbf{e}fficient solution.
Inspired from tensor decomposition methods, we find that decompose entity embedding matrix into low rank matrices can reduce more than half of the parameters while maintaining comparable performance.
To accelerate model inference, we propose a self-supervised auxiliary task, which can be seen as fine-grained entity typing. By randomly  masking and recovering entities' connected relations, the task learns the co-occurrence of entity and relations.
Utilizing the fine grained typing, we can filter unrelated entities during inference and get targets with possibly sub-linear time requirement.
Experiments on link prediction benchmarks demonstrate the proposed key capabilities. Moreover, we prove effectiveness of the proposed solution on the Open Graph Benchmark large scale challenge dataset WikiKG90Mv2 and achieve the state of the art performance.
\end{abstract}


\section{Introduction}
Knowledge graphs store structured data as the form of triples, where each triple $(h,r,t)$ represents a relation $r$ between a head entity $h$ and a tail entity $t$. Typical projects such as WordNet \cite{miller1995wordnet}, Freebase  \cite{bollacker2008freebase}, YAGO \cite{suchanek2007yago} and DBpedia \cite{lehmann2015dbpedia}, have gained widespread traction from their successful use in tasks such as question answering \cite{bordes2014open, hamilton2018embedding}, information extraction\cite{hoffmann2011knowledge}, recommender systems \cite{zhang2016collaborative} and so on. Particularly, knowledge graph embedding (KGE) methods, which embed all entities and relations into a low dimension space, play a key role for these downstream tasks. It has been an active research area for the past couple of years. Methods such as TransE \cite{bordes2013translating}, ComplEx\cite{trouillon2016complex}, RotatE\cite{sun2018rotate} provide a way to perform reasoning in knowledge graphs with simple numerical computation in continuous spaces. Although widely used in many reasoning tasks, these methods have to respond to several challenges when working with large scale KGs.

\begin{wrapfigure}[12]{r}{5cm}
    \centering
    \includegraphics[width=1.0\linewidth]{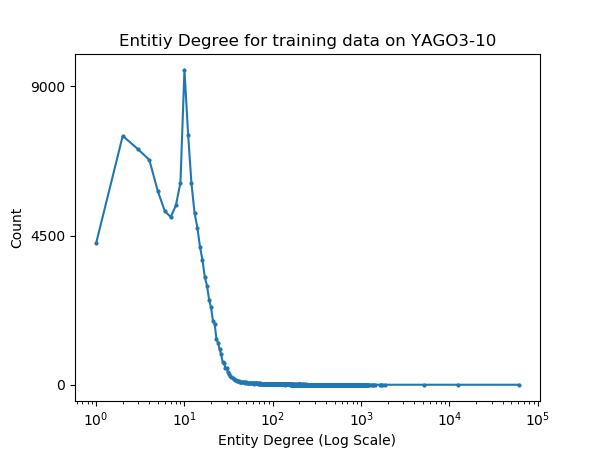}
    \caption{Entity degree distribution}
    \label{fig:long_tail}
\end{wrapfigure}

\textbf{Parameter redundancy.}
KGE methods assign the same embedding dimension for all the entities\cite{wang2017knowledge}. However, entities in KGs naturally have long-tailed distributions, where different entities have different degrees (see Figure ~\ref{fig:long_tail}). The same embedding dimension for both dense entities and sparse entities will cause either over parameterization (sparse entities) or under fitting (dense entities). On small conventional benchmark datasets, such as FB15k\cite{bordes2013translating}, YAGO3-10\cite{dettmers2018convolutional}, embedding dimensions for the state-of-the-art models can be as many as thousands \cite{sun2018rotate, ebisu2018toruse}. Furthermore, \cite{hu2020open,chao-etal-2021-pairre} show on some dataset larger dimensions can achieve better performance. Given that there are many sparse entities, especially for large scale KGs, the redundancy in the parameterization of entity embeddings will be a severe problem.
Meanwhile, the parameter issue has been a challenge for huge KGs. Since the parameters of knowledge graph embedding methods grow linearly with the number of entities and embedding dimension, increasing the dimension for huge KGs can lead to the explosion of the parameters rapidly.
With more parameters, the models consume more memory, storage and computing resources.
This motivates us to reduce parameter redundancy and improve parameter efficiency to alleviate this challenge.

\textbf{Inefficiency inference.} The inference efficiency, which has long been neglected, is also an inevitable challenge. Take the one-hop question answering reasoning task as an example. Given a test query $(?, PresidentOf,  United States)$, all the entities in KG should be traversed and sorted to find the target entities. The time complexity is $O(n + n\log(n))$, where $n$ is the size of entity set. Such a high complexity makes it difficult to scale up to large scale KGs, such as WikiKG90Mv2\cite{hu2021ogb}. One way to accelerate inference is only selecting and comparing relevant entities based on the query relation. For example, according to $PresidentOf$, we can infer the type of answers is person, or more precisely, politician. However, there are two obstacles to getting the wanted typing. The first one is KGs are incomplete, so as the entity type \cite{zhao2020connecting}. The other one is that the types are often several discrete classes \cite{yogatama2015embedding, zhouzero} such as "person" and "Location".  
We argue these coarse typing might not fine grained enough to improve inference efficiency.


To solve these two challenges, we propose PIE, a \textbf{P}arameter and \textbf{I}nference \textbf{E}fficient solution. In the field of KGE, one group of works utilize latent factorization methods \cite{nickel2011three, trouillon2016complex, amin2020lowfer}. Taking triples in KG as the form of a partially observed 3D tensor, they decompose it into a product of embedding matrices with much smaller rank, resulting in fixed-dimensional vector representations for each entity and relation.
Following this line of works, we propose low rank entity embedding (LRE). Rather than decomposing the observed 3D tensor directly, LRE decomposes the entity embedding matrix to low rank matrices.
Extensive experiments show adding LRE to existing KGE methods can reduce more than half of the  parameters while maintain comparable performance.
When the computing resource is limited, the LRE based methods show significant advantages.
To accelerate model inference, we propose an auxiliary self-supervised task, which can be seen as fine grained entity typing. Different with traditional entity typing, which defines entity types to discrete classes \cite{moon2017learning, zhao2020connecting}, we take the one-hop relation neighbor distribution (observed and unobserved) of entities as their types. This task learns the co-occurrence of entities and relations by randomly masking and recovering entities relation neighbors. Under the proposed task, we explore an inductive, entity independent and storage efficient model. Specifically, a relational message passing based graph neural network \cite{wang2021relational} is built to model entities' relation subgraphs to infer unobserved relations. With the learned fine grained entity typing, we can filter unrelated entities during inference process and get targets with possibly sub-linear time requirement.

Our key contributions are outlined as follows:
\begin{itemize}
\item[$\bullet$]
We propose utilizing tensor factorization to improve parameter efficiency of existing KGE methods, which is general and efficient;
\end{itemize}

\begin{itemize}
\item[$\bullet$]
We propose a new perspective for entity typing, which is self-supervised and fine grained. It can be used to improve  inference efficiency seamlessly. Under the proposed task, we explore an effective relational message passing based model;
\end{itemize}

\begin{itemize}
\item[$\bullet$]
On real-world extremely large benchmark, WikiKG90Mv2, the proposed solution improves both parameter and inference efficiency for existing KGE models and
achieves state-of-the-art performance under limited computing resources.
\end{itemize}

\section{Related Work}
\label{sec:headings}
\textbf{Knowledge graph embedding methods.}
Given a set of entities $E$ and relations $R$ in a KG, KGE is learned by assigning a score $s$ to a triple $(e_s, r, e_o)$:
\begin{equation}
s = f_r(e_s, e_o),\nonumber
\end{equation}
where $e_s, e_o \in E$ are the entities,  and $r \in R$ is the relation between them. The scoring function $f$ estimates the general binary tensor $T \in |E| \times |R| \times |E|$, by assigning a score of 1 to $T_{ijk}$ if relation $r_j$ exists between entities $e_s$ and $e_o$, 0 otherwise.

Based on the scoring function, previous works can be roughly classified into tensor factorization models and distance based models. Tensor factorization models use different factorization methods to decompose tensor $T$. RESCAL \cite{nickel2011three} and DistMult \cite{yang2014embedding} factorize its slices in the relation dimension. Others such as ComplEx \cite{trouillon2016complex}, SimplE \cite{kazemi2018simple} use Canonical Polyadic decomposition \cite{hitchcock1927expression} to factorize this 3D binary tensor directly.
Distance based methods \cite{bordes2013translating, wang2014knowledge, sun2018rotate, tang2019orthogonal} use additive dissimilarity scoring functions, whereby they all map entities to vector representations. All these methods do not consider the redundancy in embedding matrices, especially for entity embedding, which may be problematic for large scale KGs.

\textbf{Compression and parameter efficient models.} Inspired by the distillation technique \cite{hinton2015distilling, sanh2019distilbert},  \cite{sachan2020knowledge}  compress the trained entity matrices into discrete codes. \cite{wang2021mulde} also utilizes multiple KGE models as teachers to distill a student model, which has lower dimension.
One limitation of these works are that one or more well trained full embedding matrix is needed before distillation.
NodePiece \cite{galkin2021nodepiece} selects some of the entities as anchor entities and represents all the entities based on these anchor entities' embedding. However, it needs compute the shortest  paths from all the entities to anchor entities, which is time and resource consuming. For KG with 1M+ entities, the preprocessing step might be a computational bottleneck \footnote{As the author states, it takes 55 hours to finish the preprocess step for ogb-wiki2 (2.5M nodes and 16M edges).  https://github.com/migalkin/NodePiece/issues/1}.
Contrary to those works, the proposed method learns embeddings more efficiently.

\textbf{Entity typing.} This task aims at inferring possible missing entity type instances in KG. One way to get the type information is to infer from external text information \cite{xu2018neural}, which might not be feasible for some KGs. Others utilize embedding based models. The embedding models \cite{moon2017learning, zhao2020connecting} combine triple knowledge and entity type instances for type prediction, which require entity type instance as supervised signal.
Meanwhile, the types defined in these methods is often a small set of coarse labels.
Instead, our proposed self-supervised fine grained entity typing task can overcome these limitations, providing a new perspective to this task.
Besides, the proposed inductive relational message passing model, which aggregates the neighborhood relations, can also be a useful supplement to existing entity typing models.


\textbf{Relation prediction.} A recent line of works, such as GraIL\cite{teru2020inductive} and PathCon\cite{wang2021relational}, utilize entity-pair based subgraph and paths to predict the relation between two entities. We want to emphasize these works are different to our proposed fine grained entity typing task, where they belong to link prediction and ours are node property prediction. Besides, the fine grained entity typing is much more challenge. Because the target space for entity pair based relation prediction is certain. Due to the incomplete character of KGs, the label set can be very ambiguous for the proposed fine grained entity typing.

\section{Methodology}
\subsection{Parameter Efficient: Low Rank Entity Embedding}
\begin{figure}[t]
    \centering
    \includegraphics[width=0.7\linewidth]{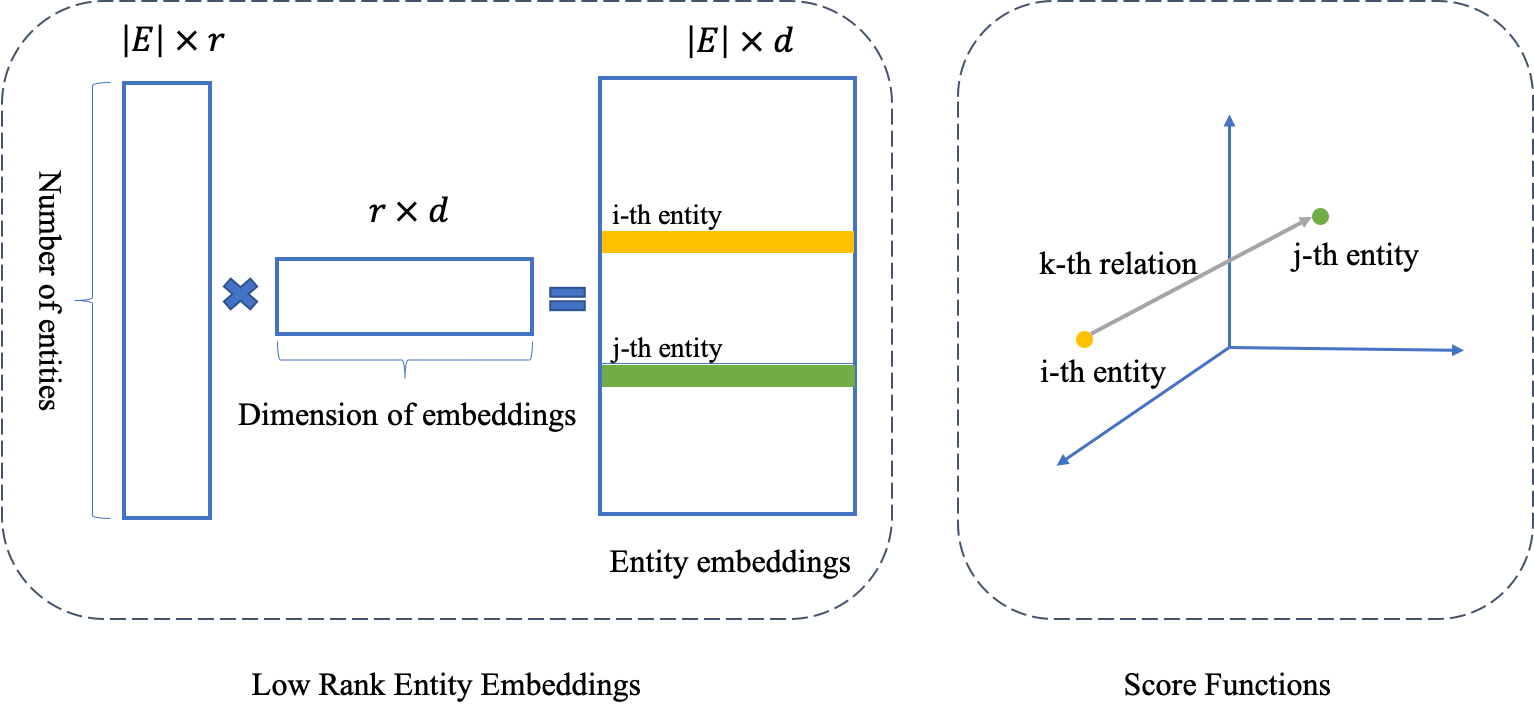}
    \caption{Low rank entity embedding for KGE methods.}
\label{fig:LRE}
\end{figure}

Conventional KGE methods assign an unique vector for each entity. Suppose the dimension of entity embedding is $d$ and the size of entity set is $|E|$. The total size of the entity embedding matrix is $d*|N|$, which is growing linearly with entity dimension and size of entity set.
Normally, assigning the same large dimension vectors for both dense entities and sparse entities will lead to parameter redundancy.
To this end, we propose LRE, which decompose entity embedding matrix into two low rank matrices.
As shown in Figure~\ref{fig:LRE}, the entity embedding matrix $Z$ is decomposed into  matrices $Z_d$ and $W$,
\begin{equation}
Z = Z_d * W,
\end{equation}
where $Z_d \in \mathcal{R}^{|E| \times r}$, $W \in \mathcal{R}^{r\times d}$ and $r < d$.
$r$ is the rank of the proposed LRE.
The matrix $Z_d$ is entity dependent and matrix $W$ is shared for all entities.
The total number of entity parameters for LRE based KGE models can be $|E| * r  + r * d$. Compared to the corresponding full entity embedding matrix based model, LRE can reduce parameters close to $|E|* (d -r)$, where the shared parameter $W$ can be neglected for large entity set. The larger the entity set, the more efficiency of LRE based model.

Although the total parameter is reduced, LRE based model has the potential to recover the performance of the larger models. In LRE, the matrix $W$ is shared for all entities. During training, the gradients of $W$ are biased to denser entities as the imbalanced distribution of entity degrees. The biased learning ensures the performance of denser entities to some extent.
While sparse entity can also learn from denser entity through the shared matrix $W$. As experiment results show with the same experiment settings, LRE can get comparable or even better performance than the 2x larger models.

LRE can be used to both tensor decomposition based models and distance based models.
When combined with tensor decomposition based models, it can be seen
as the further decomposition of the target 3D binary tensor $T$ (see Section ~\ref{sec:headings}). When getting the entity embedding, all the subsequent steps keep the same with existing KGE methods.

\textbf{Connection to NodePiece.} Both LRE and NodePiece can be seen utilizing base vocabulary to represent all entities. NodePiece randomly selects some anchor entities as vocabulary while LRE uses the shared matrix $W$.
One of the major differences is how to combine vocabulary to represent entities. NodePiece utilizes shortest path based encoding way, which is inductive but not friendly for large scale KGs\cite{galkin2021nodepiece}.
LRE utilizes a low rank learnable look-up table, which is more efficient.


\subsection{Inference Efficient: Fine Grained Typing Aware Inference}
We propose to utilize neighborhood and entity type aware inference.
Given test triple $(h, r, ?)$, we first extract neighborhoods of entity $h$. Based on query relation $r$ we can also infer the type of answer entities.
Then, we can select some candidates based on the inferred entity type. At last, the KGE models are utilized to rank all the candidates to get the answers. One of the key steps is to get the candidates based on entity type. However, KG is incomplete, including entity type. Besides, the existing entity typing may be too coarse to support efficient inference.
Thus, a fine-grained entity typing is necessary.

\begin{definition}[\textbf{Fine Grained Entity Typing}] Given a knowledge graph with entity set $E$ and relation set $R$, $\forall e \in E$, we call the distribution $p(r|e)$, $r\in R$, which measures the likelihood of relation types given a particular entity,  as fine grained entity typing of entity $e$.
\end{definition}



Once we get the fine grained entity typing, we can select candidates based on the posterior distribution, which is calculated by
\begin{equation}
p(e|r) = \frac{p(e)p(r|e)}{p(r)} \varpropto p(e)p(r|e), e \in N(e_q),
\end{equation}
where $N(e_q)$ is the neighborhood entities of query entity $e_q$.  $p(e)$ and $p(r)$ are prior distributions over entities and relations respectively, which can be calculated based on their degrees.


\textbf{Self-supervised entity typing model.}
To learn entity typing, we propose a self supervised task. Given an entity in KG, we randomly mask one of the one-hop relations and recover it by reasoning from the masked KG.
The key idea is that the missing relations can be inferred from neighborhood relations and entities. \textit{Note that the observed triples in KG can only provide partial information of the fine grained entity typing. The challenge is to infer the unobserved relation types for a particular entity.}
Apart from structure information, node attributes also provide important clues. In order to generalize KG without node attributes, we only focus on structure information in this paper.
\begin{figure}[t]
\centering
\includegraphics[width=0.9\linewidth]{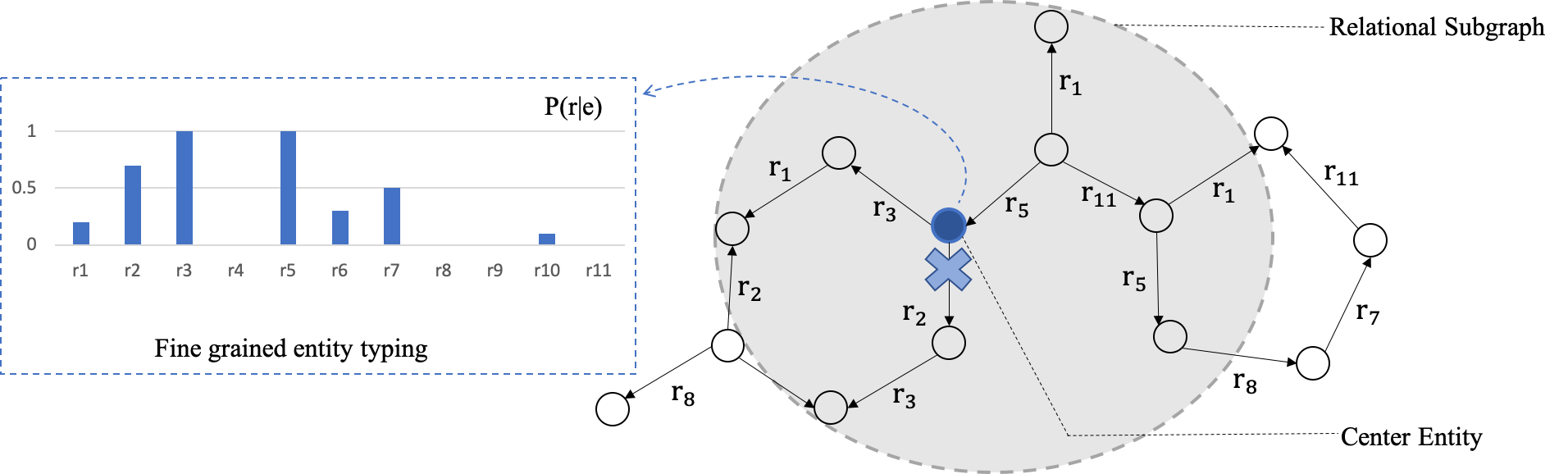}
\caption{Example of fine grained entity typing based on relational subgraph. We randomly mask one of the relation types during model learning.}
\label{fig:ERP_example}
\end{figure}

To scale up to large scale KGs efficiently, we build a lightweight inductive model.
The model is built around the Graph Neural Network (GNN) \cite{scarselli2008graph, bronstein2017geometric}.
Two modules are required: (i) relational subgraph extraction, (ii) reasoning on the relational subgraph. Example of fine grained entity typing based on relation subgraph is shown in Figure ~\ref{fig:ERP_example}.

\textbf{Relational subgraph extraction.} We assume that local graph neighborhood of a particular entity will contain the evidence needed to deduce the fine grained entity typing. As neighborhood triples with different relation types are more important than triples with the same relation, we try to keep all the relation types when we have to sample the neighbors. We also add the reverse triples with new relation types to the subgraph because edge directions for entity relation pair can have totally different semantics.

\textbf{Reasoning on the relational subgraph.}
We adopt the alternate relational message passing framework described in \cite{wang2021relational}, where node and edge representations are iteratively updated by combining their selves with aggregation of their neighbor edges' or neighbor nodes' representations.
In particular, define $s_e^i$ as the hidden state of edge $e$ in iteration $i$, and $m_v^i$ as the message stored
at node $v$ in iteration $i$. The representation of each edge is learned by:
\begin{equation}
    m_v^i =  \sum_{e \in \mathcal N(v)}{s_e^i}, \label{eq:message}
\end{equation}
\begin{equation}
    s_e^{i+1} =  \sigma([m_v^i, m_u^i, s_e^i]\cdot W^i + b^i), v,u \in \mathcal N (e),\label{eq:aggregate}
\end{equation}
where $[\cdot]$ is the concatenation function, $W^i$, $b^i$, and $\sigma(\cdot)$ are the learnable transformation matrix, bias and nonlinear activation function respectively. In our implementation, $\sigma$ is the $ReLU$ function. $s_e^0 = x_e$ is initial feature of edge $e$, which is the one hot identity vector of the relation type that $e$ belongs to. The message passing in Equation.~\ref{eq:message} and ~\ref{eq:aggregate} are repeated for
$K$ times. Inspired by the JK-connection mechanism \cite{xu2018representation}, we use the representations from all the intermittent layers and the last layer, while the final representation for the target node $t$ is represented by
\begin{equation}
    m_t = [m_t^0, m_t^1, ..., m_t^{K-1}].
\end{equation}

\textbf{Loss function.}
During learning process, the model randomly masks one relation and recover it by reasoning on the subgraph. Taking this task as a multi-classification problem, where the softmax function and cross entropy loss are utilized to maximize probability of the masked relation as other self-supervised models \cite{devlin2018bert}, is one choice. However, KG is
incomplete and the unobserved relation types except the masked relation can also be the correct labels. With larger relation set or more missing triples, the incompleteness brings new challenge for this task.
Besides, we hope the model outputs the distribution regard to all relation types rather than one specific type.
Thus, we task this task as a multi-label learning task and utilize the following loss \cite{sun2020circle}, which contains the pair wised ranking between the observed relation types and the unobserved relations. The loss is represented as
\begin{equation}
    L = \log[1 + \sum_{i \in \Omega_{obs}}{\sum_{j \in \Omega_{uno}}\exp(\gamma(s^j - s^i + m))}]  \\
    = \log[1 + \sum_{j \in \Omega_{uno}}\exp(\gamma(s^j + m)) \sum_{i \in \Omega_{obs}}\exp(\gamma(-s^i))],
\end{equation}
where $\Omega_{obs}$ and $\Omega_{uno}$ represent the observed type set and unobserved type set respectively and $|\Omega_{obs}| + |\Omega_{uno}| = |R|$. $s$ represent the logits corresponding to each relation type. $\gamma$ is a scale factor and $m$ is a margin for better separation.

\section{Experiment}
\begin{wraptable}[8]{r}{8cm}
    \centering
    \resizebox{1.0\linewidth}{!}{
    \begin{tabular}{c|c|c|c|c|c}
    \hline
    \textbf{Dataset} &\bm{$|\mathcal{R}|$} & \bm{$|\mathcal{E}|$} & \textbf{Train} & \textbf{Valid} & \textbf{Test} \\ \hline
    WikiKG90Mv2 &1,387 &91,230,610 &601,062,811 &15,000 &15,000 \\ \hline
    ogbl-wikikg2 &535 &2,500,604  &16,109,182  &429,456  &598,543 \\ \hline
    YAGO3-10 &37 &123,143 &1,079,040  &4,978 &4,982 \\ \hline
    ogbl-biokg  &51 &93,773  &4,762,678 &162,886 &162,870 \\ \hline
    FB15k &1,345 &14,951 &483,142 &50,000 &59,071  \\ \hline
    CoDEx-Large &69 &77,951 &551,193 &30,622 &30,622 \\ \hline
    \end{tabular}
    }
    \caption{\label{table:dataset} Number of entities, relations and observed triples in each split for the five utilized benchmarks.}
    \label{table:datasets}
\end{wraptable}
We first comprehensively evaluate the effectiveness of the proposed two capabilities on relative small benchmarks (Section ~\ref{sec:LRE} and Section ~\ref{sec:NTAI}). Then we report out results on one of the largest benchmark WikiKG90Mv2 (Section ~\ref{sec:90M}).

\textbf{Dataset.} We choose link prediction as evaluation task. We believe our solution can also be generalize to other reasoning tasks, such as complex question answering on incomplete KG \cite{hamilton2018embedding, saxena2021question}. The chosen benchmarks are listed in Table ~\ref{table:datasets}. Except wikiKG90Mv2, we choose ogbl-wikikg2 \cite{hu2020open}, YAGO3-10 \cite{dettmers2018convolutional},  ogbl-biokg \cite{hu2020open} and
CoDEx-Large\cite{safavi2020codex} for their entity sets are relative large. The FB15k\cite{bordes2013translating} is chosen for its large relation set, which is more difficult to infer fine grained entity typing.

\textbf{Implementation.}
For Open Graph Benchmark (OGB) related datasets, ogbl-wikikg2, ogbl-biokg and WikiKG90Mv2\footnote{The official code for wikiKG90M is utilized, which is based on DGL-KE\cite{DGL-KE}}, we utilize the official implementations. For the other two datasets, YAGO3-10 and FB15k, code from RotatE is utilized.
Since part of the OGB official implementations are based on the repository of RotatE, all the codes share the same self-adversarial negative sampling loss function \cite{sun2018rotate}.
Our code is public available\footnote{ https://github.com/alipay/Parameter\_Inference\_Efficient\_PIE}.

\subsection{Parameter Efficiency} \label{sec:LRE}

\textbf{Setup.}
We choose PairRE as baseline model. NodePiece, the entity hashing based efficient model, is also compared.
The designed experiments are mainly to prove the parameter efficiency of LRE based models rather to outperform the best existing models.

Following the state-of-the-art methods, we measure the quality of the ranking of each test triple among all possible head entity and tail entity substitutions.
Two evaluation metrics, including Mean Reciprocal Rank (MRR) and Hit ratio with cut-off value 10, are utilized.
For experiments on ogbl-wikikg2 and ogbl-biokg, we follow the evaluation protocol of these two benchmarks \cite{hu2020open}.
We also report the total parameters for all models and the required hardware for larger dataset.

\begin{table*}[hp]
\begin{center}
\resizebox{1.0\textwidth}{!}{
\begin{tabular}{c|cccccc|ccccc}
\hline
- &\multicolumn{6}{c|}{ogbl-wikikg2} & \multicolumn{5}{c}{ogbl-biokg} \\ \hline
\textbf{Model} &$\#$Dim &$\#$LRE\_Rank &Test MRR &Valid MRR &$\#$Param(M) &GPU &$\#$Dim  &$\#$LRE\_Rank &Test MRR &Valid MRR &$\#$Param(M) \\ \hline
AutoSF\cite{zhang2020autosf} &200 &- &0.5458 &0.5510 &500.2 &45G &1000 &-
&0.8309 &0.8317 &93.8 \\
AutoSF+NodePiece\cite{galkin2021nodepiece} &200 &- &\textbf{0.5703} &\textbf{0.5806} &6.9 &32G &- &- &- &- &- \\
\bottomrule \bottomrule
PairRE\cite{chao-etal-2021-pairre} &200 &- &$0.5289$ &${0.5529}$ &500.3 &16G &2000 &- &${0.8164}$ &${0.8172}$ &187.8 \\
PairRE &400 &- &$0.5805$ &${0.5529}$ &1,000.7 &32G &- &- &- &- &- \\
PairRE+LRE &700 &200 &$0.5836$ &$0.5861$ &501.0 &16G &2000 &800 &$0.8244$ &$0.8250$ &76.6 \\
PairRE+LRE &5000 &200 &$\textbf{0.5970}$ &$\textbf{0.6033}$ &506.5 &24G &5000 &2000 &$\textbf{0.8360}$ &$\textbf{0.8367}$ &198.1 \\ \hline
\end{tabular}
}
\end{center}
\caption{\label{table:ogbl} Link prediction results on ogbl-wikikg2 and ogbl-biokg.
Best results for each group are in bold.
Each experiment is run ten times. We do not add the standard deviations for limited space.}
\end{table*}

\begin{table*}[hp]
\begin{center}
\resizebox{1.0\textwidth}{!}{
\begin{tabular}{c|ccccc|ccccc}
\hline
- &\multicolumn{5}{c|}{YAGO3-10} & \multicolumn{5}{c}{CoDEx-Large} \\ \hline
\textbf{Model} &$\#$Dim &$\#$LRE\_Rank &MRR &Hit@10 &$\#$Param(M) &$\#$Dim  &$\#$LRE\_Rank &MRR &Hit@10 &$\#$Param(M) \\ \hline
RotatE\cite{sun2018rotate} &500 &- &\textbf{0.495} &\textbf{0.670} &123 &500 &- &\textbf{0.258} &\textbf{0.387} &77  \\
RotatE+NodePiece\cite{galkin2021nodepiece} &500 &- &0.247 &0.488 &4.1 &500 &- &0.190 &0.313 &3.6 \\
\bottomrule \bottomrule
PairRE\cite{chao-etal-2021-pairre} &700 &- &\textbf{0.540} &\textbf{0.693} &86.3 &1000  &- &\textbf{0.290} &\textbf{0.415} &78.1 \\
PairRE+LRE &1000 &30 &$0.284$ &$0.472$ &3.8  &1000 &30 &$0.234$ &$0.375$ &2.5
 \\ \hline
\end{tabular}
}
\end{center}
\caption{\label{table:yago} Link prediction results on YAGO3-10 and CoDEx-Large.}
\end{table*}


\begin{wrapfigure}[16]{r}{6.5cm}
    \centering
    \includegraphics[width=1.0\linewidth]{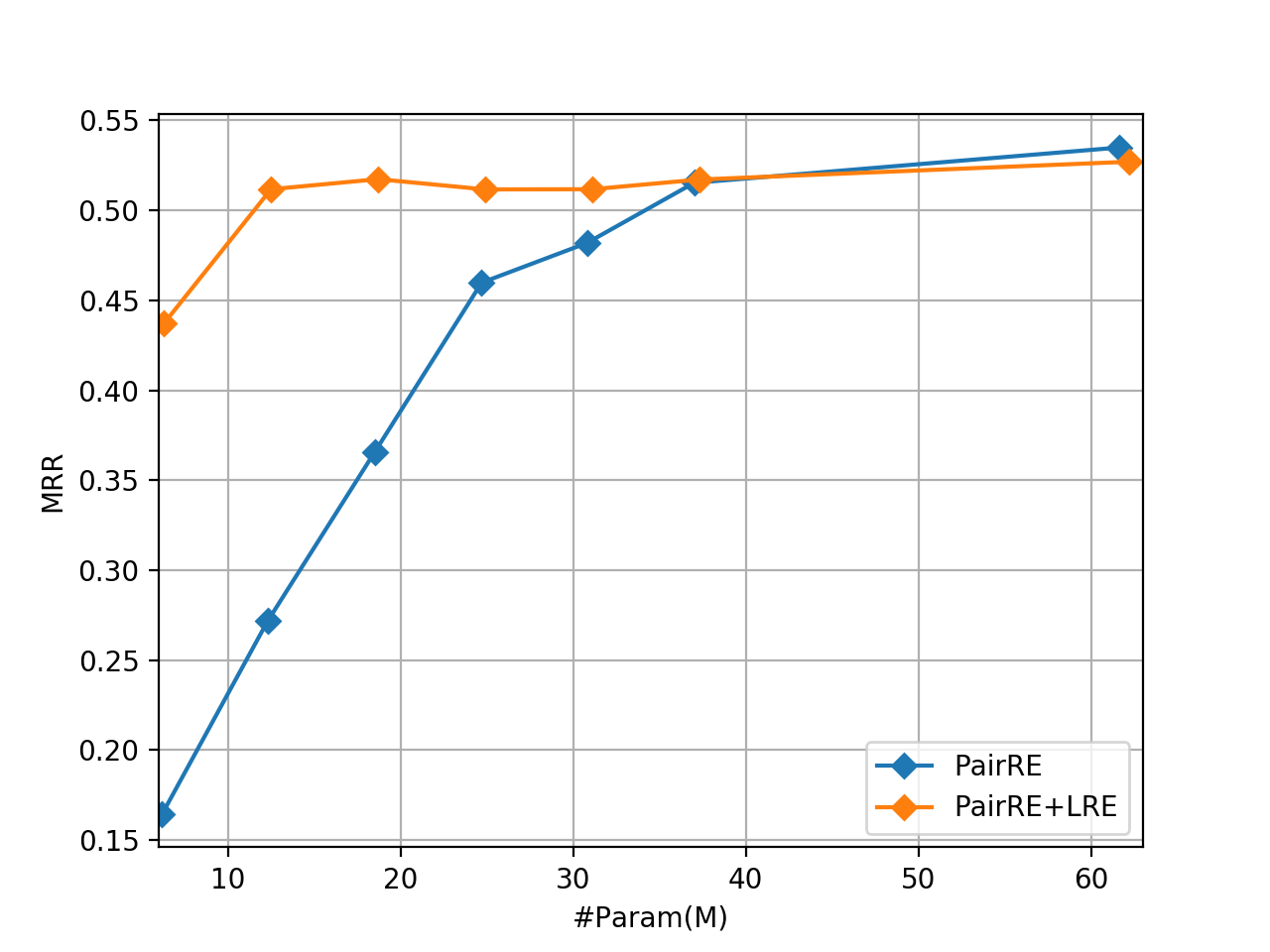}
    \captionof{figure}{Parameter efficiency comparison on YAGO3-10 dataset.}
    \label{fig:PairRE_YAGO}
\end{wrapfigure}


\textbf{Results.} Comparisons for ogbl-wikikg2 and ogbl-biokg are shown in Table ~\ref{table:ogbl}. All the models are in the same experiment settings and implementation.
The results shows larger dimension can improve the Test MRR significantly on ogbl-wikikg2. While the LRE based models can boost the performance further with the same budget of parameters and computing resource. The same phenomenon can also be observed on ogbl-biokg. The LRE based models can improve test MRR with 6.9\% and 1.96\% respectively. Results for YAGO3-10 and CoDEx-Large are shown in Table ~\ref{table:yago}.
When compared to NodePiece based model, on ogbl-wikikg2, LRE based model consumes more parameters to outperform their performance. However, as stated before, LRE based models can avoid the long preprocess time of NodePiece based methods. On YAGO3-10 and CoDEx-Large, the LRE based model can outperform NodePiece based model with smaller budget of parameters (see Table ~\ref{table:yago}).
All these experiments show LRE based models is advantageous when meet with limited resource and large scale datasets.

We further study the influence of the parameter size to the performance (Figure ~\ref{fig:PairRE_YAGO}). On YAGO3-10, the performances of PairRE and PairRE+LRE increase with the increase of parameter size.
The dimension of PairRE+LRE is set to 1000 and the LRE rank is increasing from 50 to 300. While the dimension of PairRE is increased from 50 to 500.
The performance of PairRE+LRE saturates at 20M, while PairRE at 60M. When the size of parameter is less than 40M, PairRE+LRE shows much better performances.
Particularly, LRE powered models can achieve comparable performances compared to 3x larger models (LRE+PairRE@12M VS PairRE@38M).
This indicates the proposed LRE can improve the parameter efficiency for KGE methods.


\subsection{Inference Efficiency} \label{sec:NTAI}
\begin{wraptable}[9]{r}{6.5cm}
\begin{center}
\resizebox{1.0\linewidth}{!}{
\begin{tabular}{c|cc|cc}
\hline
- &\multicolumn{2}{c|}{YAGO3-10} & \multicolumn{2}{c}{FB15k} \\ \hline
Loss &MRR  &Hit@5 &MRR &Hit@5 \\ \hline
Softmax  &0.9765 &\textbf{0.9972} &0.7005 &0.8495 \\
Ranking loss &\textbf{0.9898} &0.9971 &\textbf{0.9017} &\textbf{0.9391} \\ \hline
\end{tabular}
}
\end{center}
\caption{\label{table:yago_relation_prediction} Evaluations for the entity typing model on YAGO3-10 and FB15k.}
\end{wraptable}
\textbf{Setup.} We first report the proposed entity typing model performances on YAGO3-10 and FB15k. The comparison between softmax based loss and the proposed ranking loss is added as we find loss function is a vital step in our experiments.
When evaluating the entity typing model, we follow the data splits of the conventional link prediction task. For a given triple $(h, r, t)$ in the test or valid set, we rank the ground truth relation type $r$ against all other  relation types based on the relation subgraph of entity $h$.
We also use MRR and Hit@5 as evaluation metrics. The rankings are computed after removing all the other observed relation types that appear in training set.

\begin{wrapfigure}[19]{r}{6.5cm}
    \centering
    \includegraphics[width=1.0\linewidth]{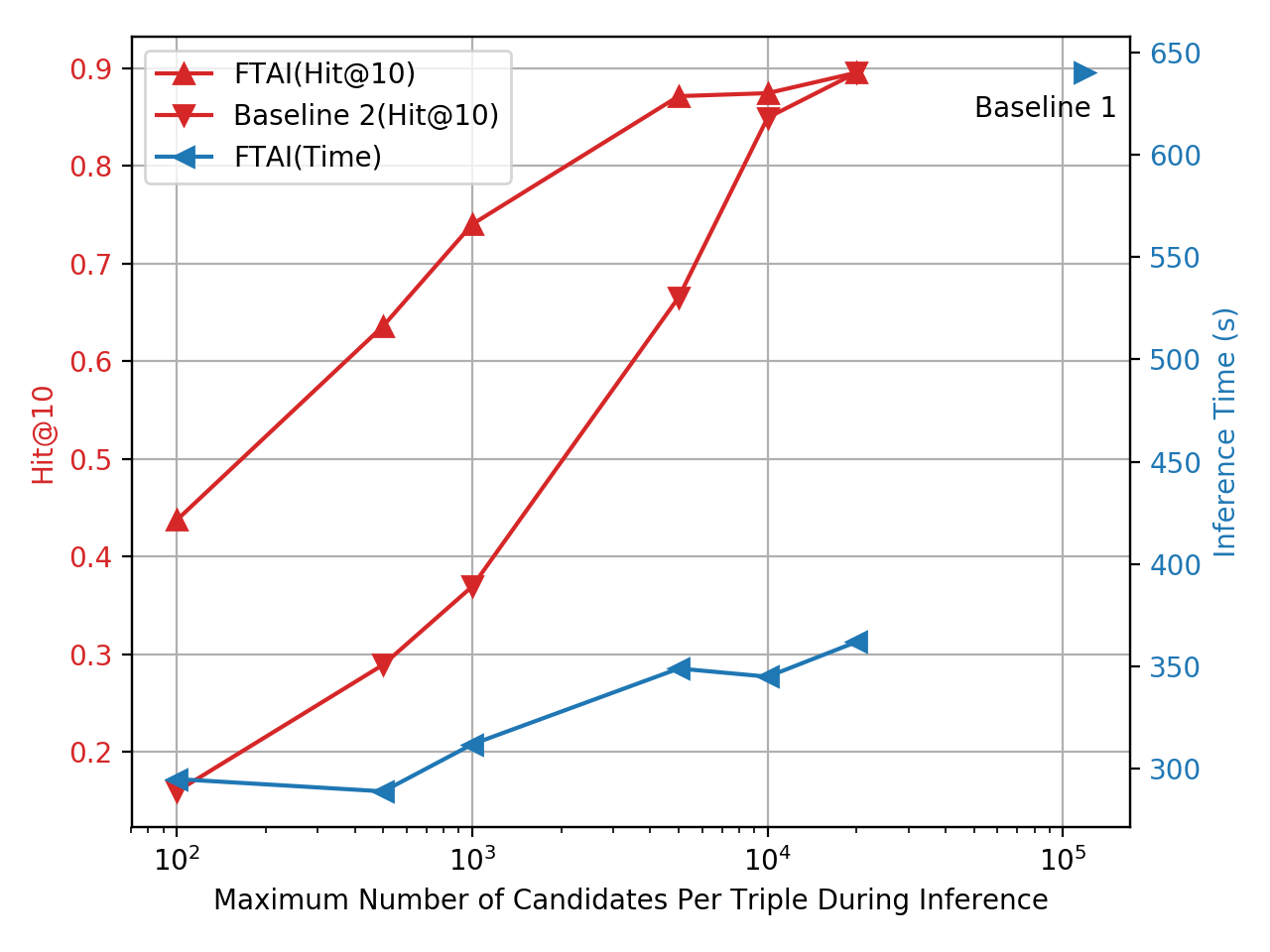}
    \caption{Influence of the candidate number per test triple of FTAI on test MRR and inference time on FB15k for the PairRE model. Baseline 1 is scoring all entities. Baseline 2 is the proposed FTAI without use entity typing result. }
    \label{fig:FB15k_time_mrr_candidate_num}
\end{wrapfigure}
Then two baselines are compared to the fine grained typing model powered inference. We name the proposed inference method as \textbf{F}ine-grained entity \textbf{T}yping \textbf{A}ware \textbf{I}nference (FTAI). Baseline one is the conventional inference, which is scoring and ranking all the entities.
Baseline two is the neighborhood based inference. The difference between FTAI and baseline two is that only entity degree is used to select candidates from neighborhood entities for baseline two. The second baseline can be seen as the ablation analysis for the proposed entity typing model. Two evaluation metrics are utilized, inference time and Hit@10.


\textbf{Results.} Table ~\ref{table:yago_relation_prediction} shows the evaluation results of the entity typing models. Clearly, the FB15k dataset is more challenge as there is much larger relation sets compared to YAGO3-10 (1,315 vs 37). The table also shows pronounced gap between softmax loss and the proposed ranking loss on FB15k. One of the reasons  might be the incompleteness of FB15k is more severe. Under this situation, the ranking loss based model can largely outperform softmax based model.

Incorporate the entity typing model, the proposed FTAI shows clear advantages compared to these two baselines (See Figure \ref{fig:FB15k_time_mrr_candidate_num}). Compared to the conventional inference, FTAI can achieve 1.8x speedup and the same test Hit@10 performance when at most 20k candidates are selected for each test triple. Ablation analysis shows for both FTAI and baseline two, with larger candidate set, the better performance and the slower inference. The results also show the entity typing model enables a much large recall compared with baseline two, which proves the crucial function of the proposed entity typing model.

\subsection{Results on WikiKG90Mv2} \label{sec:90M}
\textbf{Setup.} We follow the official evaluation of WikiKG90Mv2. That is, for each $(h, r, ?)$, the model is asked to predict the top 10 tail entities that are most likely to be positive.
Since traversal all the 90 million entities is impractical during inference, the organizers of this dataset provide at most 20,000 tail candidates for each test triple.
They pick tail candidates $t$ by choosing the entities with at least one triple $(\_, r, t)$ on the KG, and sort all the tails by its degree. We take these candidates as baseline candidates.
In accordance with baseline models, we also pick at most 20k candidates for FTAI.

\textbf{Results.}
The reported results (Table ~\ref{table:90mv2}) show competitive performances of PIE powered ComplEx and TransE models.
We tune the baseline models carefully and get much better performances compared to the official results. With the same hyper-parameter settings and parameter budget, LRE can improve ComplEx and TransE 0.56\% and 0.75\% respectively. Utilizing the fine grained entity typing model, we can improve ComplEx and TransE 0.97\% and 2.47\% respectively. This is because the candidates predicted by our entity typing model have much higher recall.
As Table \ref{table:90M_recall} shows the candidates output by FTAI have a much higher recall than the baseline candidates.
When combine TransE and the proposed PIE, we achieve state-of-the-art performance, with 2.82\% improvement compared to the entity attribute powered model.

\begin{table}[ht]
\begin{center}
\resizebox{0.7\textwidth}{!}{
\begin{tabular}{c|ccccc} \hline
\textbf{Model} &$\#$Dim &$\#$LRE\_Rank &Test MRR &Valid MRR &$\#$Param(B) \\ \hline
TransE-Concat\cite{hu2021ogb}$\Diamond$ &200 &- &0.1761 &0.2060 &18.2 \\
\bottomrule \bottomrule
ComplEx\cite{hu2021ogb} &100 &- &0.0985  &0.1150 &18.2 \\  \hline
ComplEx(ours) &100 &- &- &0.1795 &18.2 \\
+LRE &400 &200 &- &0.1851 &18.2 \\
+FTAI &100 &- &- &0.1892 &18.2 \\
+LRE+FTAI &400 &200 &- &\textbf{0.1919} &18.2 \\
\bottomrule \bottomrule
TransE\cite{hu2021ogb} &200 &- &0.0824 &0.1103 &18.2  \\  \hline
TransE(ours) &200 &- &- &0.2049 &18.2 \\
+LRE &600 &200 &- &0.2124  &18.2 \\
+FTAI &200 &- &- &0.2296  &18.2 \\
+LRE+FTAI &600 &100 &- &0.2247 &9.1 \\
+LRE+FTAI &600 &200 &\textbf{0.1883} &\textbf{0.2342} &18.2 \\
\hline
\end{tabular}
}
\end{center}
\caption{\label{table:90mv2} Link prediction results on WikiKG90Mv2.$\Diamond$ means the entity attribute is utilized.}
\end{table}

\begin{wraptable}[7]{r}{5cm}
\begin{center}
\resizebox{1.0\linewidth}{!}{
\begin{tabular}{c|c}
\hline
Model &Recall@20k \\ \hline
Baseline\cite{hu2021ogb} &0.5770 \\
FTAI &\textbf{0.6695}  \\ \hline
\end{tabular}
}
\end{center}
\caption{\label{table:90M_recall} Evaluations for entity typing model on WikiKG90Mv2 valid set.}
\end{wraptable}
Ablation analysis result of the proposed FTAI is shown in Table ~\ref{table:90M_recall}. With the same candidate number, the recall of FTAI is higher than the baseline candidates with close to 10\%, which proves the necessity and the effectiveness of the fine grained typing model.
The evaluation metrics for entity typing model on this dataset is 65\% for MRR and 72\% for Hit@5. Compared to results on FB15k, there is still large room for further improvement. We leave this as future work.

\section{Conclusion and Future Work}
With the explosion of data, the emerging large scale KGs bring new challenges for KGE based reasoning. We explored the parameter redundancy and inefficiency inference. For parameter redundancy, we propose tensor factorization based method LRE, which decomposes the entity embedding matrix.
Experiment results show LRE based model can approximate the performance of full rank entity matrix based models, which saves more than 50\% parameters.
For inference optimization, we proposed a fine grained entity typing model.
Without extra annotation, we learn the concurrence distribution of entity and relation with a lightweight and inductive model. Utilizing the learned entity typing, the full entity set traversing is avoid during inference. Experiment results on FB15k shows we achieve a 1.8x speedup over the  full entity set traversing inference. We also verify the proposed solution on large scale dataset WikiKG90Mv2 and achieve state of the art performance.

The deep models have shown tremendous successes \cite{goodfellow2016deep}.
Compared to models from computer vision \cite{he2016deep} and NLP\cite{devlin2018bert}, where the layer-wise hierarchy representations are widely used, KGE models are shallow.
The proposed LRE can be seen as an exploration from shallow to deep, where the large look-up table is replaced with a smaller table plus a linear layer.  For future work, we will try more "decomposition" for the look-up table with techniques such as residual connection\cite{he2016deep} and auxiliary losses in intermediate layers\cite{szegedy2015going}.
The incompleteness of KGs also leads to challenge for the fine grained entity typing. A framework that unifies KG reasoning and fine grained entity typing should boot performances of these two tasks. We will leave these for future works. 

\bibliographystyle{unsrtnat}
\bibliography{template}

\end{document}